# Dependency resolution and semantic mining using Tree Adjoining Grammars for Tamil Language

**Vijay Krishna Menon**
Centre for Excellence in Computational
Engineering and Networking (CEN),
Amrita School of Engineering,
Amrita Vishwa Vidyapeetham, Coimbatore.
`m_vijaykrishna@cb.amrita.edu`

**Rajendran S**
Centre for Excellence in Computational
Engineering and Networking (CEN),
Amrita School of Engineering,
Amrita Vishwa Vidyapeetham, Coimbatore,
`rajushush@gmail.com`

**Anand Kumar M**
Centre for Excellence in Computational
Engineering and Networking (CEN)
Amrita School of Engineering
Amrita Vishwa Vidyapeetham, Coimbatore
`m_anandkumar@cb.amrita.edu`

**Soman K P**
Centre for Excellence in Computational
Engineering and Networking (CEN)
Amrita School of Engineering
Amrita Vishwa Vidyapeetham, Coimbatore
`kp_soman@amrita.edu`

**Abstract**

Tree adjoining grammars (TAGs) provide an ample tool to capture syntax of many Indian languages. Tamil represents a special challenge to computational formalisms as it has extensive agglutinative morphology and a comparatively difficult argument structure. Modelling Tamil syntax and morphology using TAG is an interesting problem which has not been in focus even though TAGs are over 4 decades old, since its inception. Our research with Tamil TAGs have shown us that we can not only represent syntax of the language, but to an extent mine out semantics through dependency resolution of the sentence. But in order to demonstrate this phenomenal property, we need to parse Tamil language sentences using TAGs we have built and through parsing obtain a derivation we could use to resolve dependencies, thus proving the semantic property. We use an in-house developed pseudo lexical TAG chart parser; algorithm given by Schabes and Joshi (1988), for generating derivations of sentences. We do not use any statistics to rank out ambiguous derivations but rather use all of them to understand the mentioned semantic relation with in TAGs for Tamil. We shall also present a brief parser analysis for the completeness of our discussions.

## 1. Introduction

TAGs were proposed for language models earlier by Vijay Shankar and Aravind Joshi in (Vijay-Shankar and Joshi, 1985). Unlike the Chomskian formalisms, the elementary objects manipulated by TAG are trees; structured objects and not strings. Such structured formalisms have properties that relate directly to strong generative capacity (structure descriptions), which is linguistically more relevant than string sets (weak generative capacity). So we call TAGs as a tree generating system rather than a string generating system. The set of all trees derived in a TAG constitute the object language. Hence, in order to describe the derivation of a tree in the object language, we will need to know about 'derivation trees'. The derivation trees are important in both syntactic and semantic senses. TAGs also have some interesting linguistic properties. Lexicalization is one of the key motivations for the study of TAGs, both linguistic and formal. The lexical phenomena now explain many linguistic theories previously thought to be purely syntactic. So the information in lexicons, have increased both in amount and complexity. From the formal perspective, lexicalization allows us to associate every elementary structure (trees) with a lexicon (any word). The famous Greibach Normal Form (also called Chomsky Normal Form or CNF) for CFGs is a kind of lexicalization. However it is a weak lexicalization, as the structure of the original grammar is not preserved and all rules cannot be lexicalised. Thus TAGs provide an edge to this errand over conventional CFGs.

TAGs were introduced by Joshi et al. (1975) and later Joshi (1985). It is known that tree adjoining languages (TALs) generate some strictly context sensitive languages and fall in the class of the so called 'mildly context sensitive' languages (Joshi et al, 1991). TALs properly contain context-free languages and are properly contained by indexed languages. A tree-adjoining grammar (TAG), G consists of a quintuple ($\sum$, NT, I, A, S) where

i. $\sum$ is a finite set of terminal symbols. NT is a finite set of non-terminal symbols such that ($\sum \cap$ NT) = $\phi$.
ii. S is a Sentential symbol such that S $\epsilon$ NT.
iii. I is a finite set of trees called initial trees, with the following properties
    a. Interior nodes are labelled by non-terminal symbols;
    b. The nodes on the frontier of all initial trees are labelled by terminals or non-terminals; non-terminals symbols on the frontier of any tree in I are marked for substitution which, by convention is a down arrow ($\downarrow$);
iv. A is a finite set of trees called auxiliary trees, with the following properties
    a. Interior nodes are labelled by non-terminal symbols;
    b. The nodes on the frontier of auxiliary trees are labelled by terminal symbols or non-terminal symbols. Non-terminal symbol on the frontier of trees in A are marked for substitution except for one node, called the foot node; by convention this is marked with an asterisk(*); the label of the foot node must be identical to the root node.

In lexicalised TAG, at least one frontier node must be labelled with a terminal symbol (the anchor) in all initial and auxiliary trees. The set I U A is called the set of elementary trees. If an elementary tree has its root labelled by non-terminal X, then it is called an X-type elementary tree.

A tree built by combining the elementary trees is called derived tree or parse tree. We will now have to understand how the combinations of trees happen as to make a derived tree. There are 2 major composition operations adjoining and substitution.

Adjoining (or adjunction, as it is alternately referred) builds a new tree from an auxiliary tree β and a tree α (α is any tree initial auxiliary or derived). Let 'α' be a tree containing a non-substitution node labeled by X. The resulting tree, γ, obtained by adjoining β to α at node n is structured as:

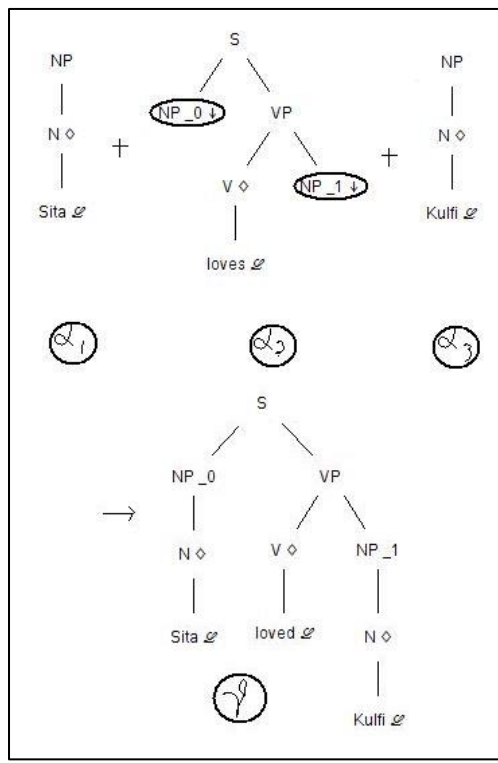

Figure 1: Susbstitution of trees in TAG

- The sub-tree of α with root n is displaced by β, along with its root node n.
- The displace sub tree of α will attach itself to β, replacing the foot node of β.

Substitution takes place only on non-terminal nodes in the frontier of a tree. Unlike normal adjunctions, substitutions are mandatory if the node is marked for it with a down arrow as explained above. When a node, say n, is substituted, the entire node is replaced by the initial tree that is substituted. Only initial trees or its derivatives may be used for substitution. By definition adjunctions on any node marked for substitution is not permitted. But adjunctions are possible on the root nodes of the trees already substituted replacing the marked node. This is illustrated in Fig 2 with a set of three initial trees. Substitution extents the targeted leaf node to complete a construct that requires addition of a single substring.

When TAG grammar yields (generates) derived trees by derivation, the information to trace the history of such combination is not given. Unlike CFGs, the derived tree does not contain information as to which basic rules (in our case, elementary trees) were used to construct it.

Hence we require a new object that gives us information regarding all operations and elementary trees used to build a derived tree. This structured object is called a derivation tree. It uniquely specifies what operation was used to combine which particular trees. Both adjunctions and substitutions are considered for derivation.

## 2. Derivation Structures in TAG

Consider the example sentence "Yesterday a man saw Mary". This example has been adopted from Joshi and Schabes (1997). Fig 3 illustrates the derived tree for the above English sentence. But this tree does not give any relevant information regarding how it can be constructed. For this we define the derivation tree for the same sentence. Refer to Fig 4 where the necessary elementary trees required to derive the α5 has been illustrated. Note that α trees are initial trees and the β ones are auxiliary. This convention will be prevailing throughout this paper whenever referring to TAG trees.

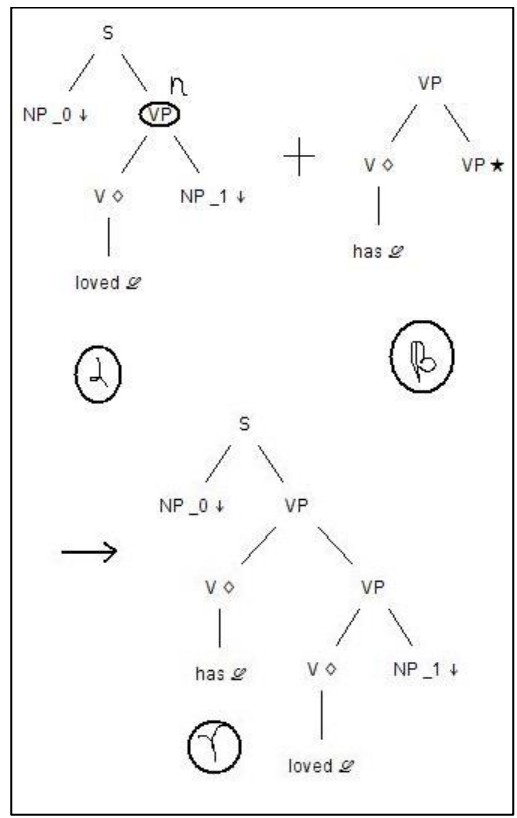

Figure 2: Adjoining of elementrt trees

Now the derivation tree for this example is shown in Fig 3. Along with exemplifying the process of building a derivation we also show how a proper lexicalization of TAG is achieved. All the elementary trees in the Fig 3. are properly and completely lexicalised with every elementary tree mapped to at least one lexicon. So every tree will have at least one anchor node.

The roots of all derivation trees are labelled by the name of an S-type initial tree. All child nodes are labelled by auxiliary trees which adjoined or initial trees which are substituted. The notion of tree address is used here to indicate where the composition happened. This will uniquely identify a node in a given tree. This address is referred to as the Gorn index; used for multiple array of purposes and is specifically important from an implementation point of view.

The Gorn index system starts with index 0 for the root node. For the 1st level children the numbering starts with 0.1 (or just 1) for the leftmost and increasing towards the right. For the 2nd level children say the child of the second leftmost child will be given 0.2.1 (or just 2.1) and so on. The system is simple and intuitive. Now if an adjunction takes place at this node of the tree, the derivation tree node labelled with the adjoining auxiliary tree will also carry the Gorn index 0.2.1, so we know exactly where the adjunction or substitution has occurred.

Fig 3 depicts the derivation and elementary trees for the mensioned example. Note that $α_{saw}$ is an S-type initial tree; most verb initial trees are expected to be so. Now the node αman (1) indicates a substitution of this tree at node 0.1 of $α_{saw}$. In a deeper sense it means this tree replaced the node indexed 0.1 in tree $α_{saw}$.

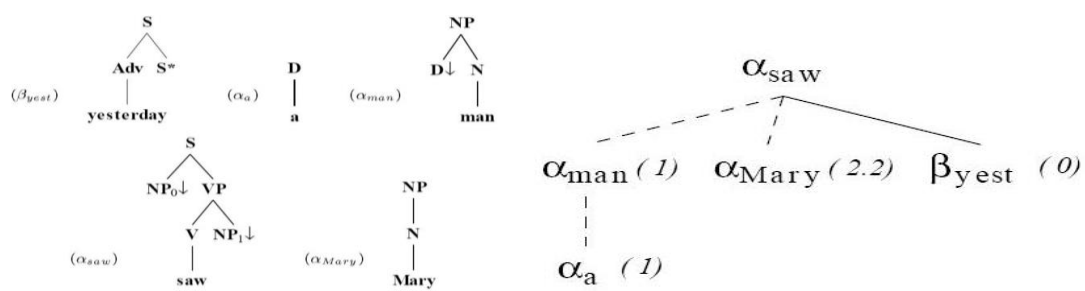

Figure 3: Derivation tree(on the right) and elementary lexicalised trees

The case with α_Mary is no different, except that it is substituted at for node 0.2.2. But β_yesterday is an auxiliary tree and is adjoined at the root node of αsaw as it contains the Gorn index pointing to the root. The main idea here is the Gorn indices given in a derivation tree's node, points to an address in its parent node's tree where the substitution or adjunction has been done. Further it also demonstrates how lower composition happens, like α_a substituted on α_man. Unlike as represented, substitutions need not be discriminated with dotted lines alone. The target node tree can solve the conflict by its type as in initial or auxiliary. Another counter intuitive fact is that adjoining happens even at the root node. But controlling adjunctions will help us control the grammars generative ability and restrict the constructs it creates. So every node in the derivation tree will have distinct indices for a given parent node. This way of representing derivation not only captures the syntactic structure of the target tree but also contains semantic dependencies. This has been demonstrated by Joshi and Rambow (1997); they were the first to investigate this property for TAG derivations. Later, Joshi and Rambow (2003) gave a dependency grammar based on TAG formalism. However we shall give a different picture of the same idea here. To illustrate this let us isolate the basic words of the above given example itself. Before we go into detail of this we will need to define dependency functions of each word with respect to the parts of speech (POS) of each word. Consider initially the verb saw. Now 'saw' is a transitive verb[1], so it will have dependencies in 2 ways, one with its subject and the other with the object. Hence the dependency function will look like this (basic argument structure).

$$f[d](saw) = verb[t](sub, obj) = verb[t](man, Mary)$$

This show the dependencies of the transitive verb saw to depend on the subject as to who or what saw to the object as to saw whom or what. Logically this function looks like this for saw.

$$< Who > saw < Whom >$$

This is exactly what we get in the derivation; "man saw Mary" giving us the dependency function for saw to be saw (Man, Mary). All the other words will have dependencies too as well. As for the Noun man the function is different and addresses the number or specificity. That means that nouns have articles or adjectives that describe them. This is their dependency. The above derivation also gives man(a) which is the dependency function for the word. The dependencies of a word can be easily found from the children of the given node in a derivation tree.

From the above insight, we must gather that saw in this example is not just transitive. That is to say it has a subject, an objects and an adverb. Thus the definition of the function should be having an extra parameter, one that specifies time in this case hence we have saw (Man, Mary, Yesterday). This property of TAG derivation greatly helps for representation of agglutinative languages, where the verbal inflection will depend on its subject or object or both. Subject verb agreements are crucial especially in Indian languages.

## 3. Tamil TAG and Derivations in Tamil

Tamil is a morph rich language, so to do pure syntax based dependency mining from it we will need to set aside the morphological considerations for the while and focus on the syntactic and psycho syntactic models. We have hand developed a Tamil TAG. Though its scope is quite restricted and tested mainly on tourism and health based corpora, it is effective enough for text book class sentences. Since such sentences only have limited or light dependencies, it might just prove to be insufficient for detailed analysis, how ever our attempt can be considered a step one into TAG based semantic analysis for Tamil language.

The Tamil TAGs were mainly created as part of a Machine Translation project, using synchronous TAGs. So these grammar trees are synchronised over a subset of XTAG English grammar. Efforts are being made for this to be expanded to a comprehensive grammar not just limited to tamil. Unlike general XTAG trees we have designed single anchor trees; a grammar tree can be lexicalised only with one lexicon. This way we maintain a one to one relation between lexicons and derivation nodes so that the node represents only the dependancy relation of that perticular lexicon. We will try to explain this through a set of examples in Tamil. Also note that we donot do any kind of

---
[1] Verbs that require a subject and an object of action are transitive verbs.

statistical parsing or context based ranking of parses over the sentence. To fully observe the dependencies, all syntactically ambiguous parses are needed, so as to obtain different points of views and preserve the natural ambiguity. Before we observe the parses we need to describe the main aspects of the grammar. We have tried and captured the following few main constructs of Tamil

1. Noun, Postpositions (morphemes), Conjuctions, Adverbs
2. Reccursive ajdectives
3. Basic Clefts (If clefts in a limited way)
4. Transitive Verb
5. Intransitive Verb
6. Ergative Verb
7. PP complement
8. PP small Clause
9. Sentential Complement
10. Sentential Subject

These constructs are reprsented using elementary trees and auxiliary trees as was seen fit liguistically and by ease of parsing. The Parser is a multithreaded java implementation of the 'Earley Type TAG parsing' algorithm by Shabes and Joshi (1987). It generates both parse trees and derivation trees over each and every ambiguous parse it can find from the grammar provided. Fig 4 demostrates the the Tamil TAG trees as redered by our viewer. The anotations for the nodes are consistent with the XTAG conventions except that all trees have sigle anchor node that houses the POS category the tree belongs to. The figure eveidently shows a mapping between the English and Tamil trees as earlier mentioned.

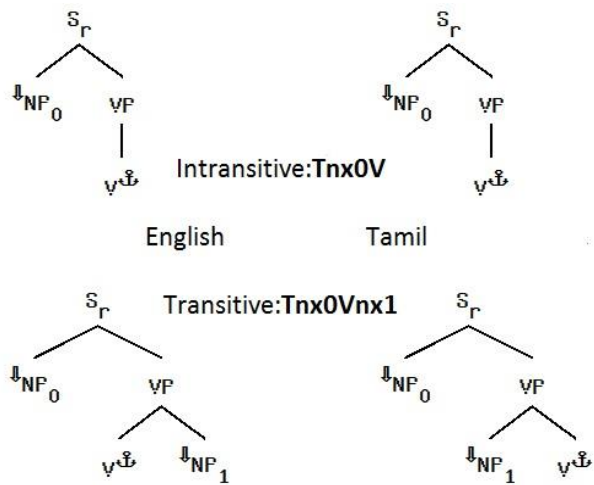

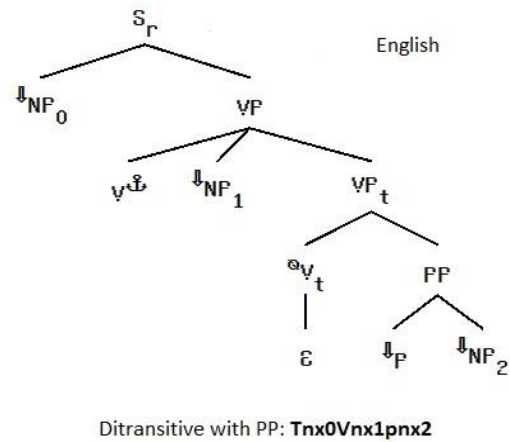

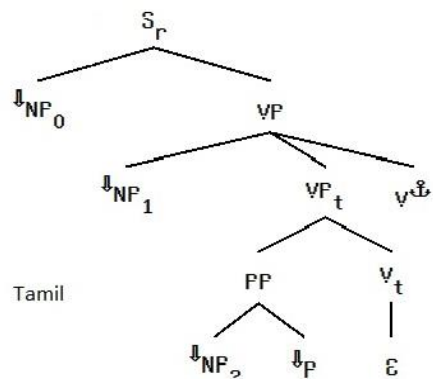

This however does not diminish the generative capacity of the Tamil grammar to independantly parse and generate derivation trees on its own accord. Positively it helps to allign the dependacy with english like dependency and base it on the stanford dependency set of 40 major semantic relations.

We however will deal with just one or two cases as a proof of concept. To mine out the relations, we introduc argument operators on certain lexical items, that will hypothetically give us the semantic argumets of the item's roll in the sentence from a TAG derivation. Each derivation tree is defined recursively to yeild the arguments when an operator operates over it.

The dependencies that we mine here as part of a miniature experiment are the following stanford depandancies

1. Nominal Subject (nsub)
2. Direct Object (dobj)
3. Root (root)

## 4. Tamil parse specifics and examples

The grammar used by the parser for Tamil cotains over 120 trees conrectly and are regularly pruned to reduce cross ambiguities. We have 25 initial and 95 auxiliary trees. Together they address most constructs of simple and direct sentences.

The parser accepts Parts of speech tagged sentences using the Penn Tagset for the same. If the sentence is with in the construct range of the grammar, the parser immediate returns TAG derivations from which a derived trees can be easily constructed. As mentioned before we are not currently deling with morph analysis just to keep our focus on grammar and parsing. In the examples here the words are mostly surface forms with some chucks in it. Our secondary objective is to prove the conformity of TAG syntax for Tamil in a broader sense. Two examples of the parse instances are illustrated bellow. The Tamil sentences has been Romanised for the sake of linguistic verification.

***Example 1***: *NiyUyArkkil naṭaipeRRa yu.Es-OpaN-Aṇkaḷ-iraṭṭaiyar iRutip-pOṭṭiyil liyAṇṭar-payas-jOṭi veRRi peRRu paṭṭattaik kaippaRRiyatu*

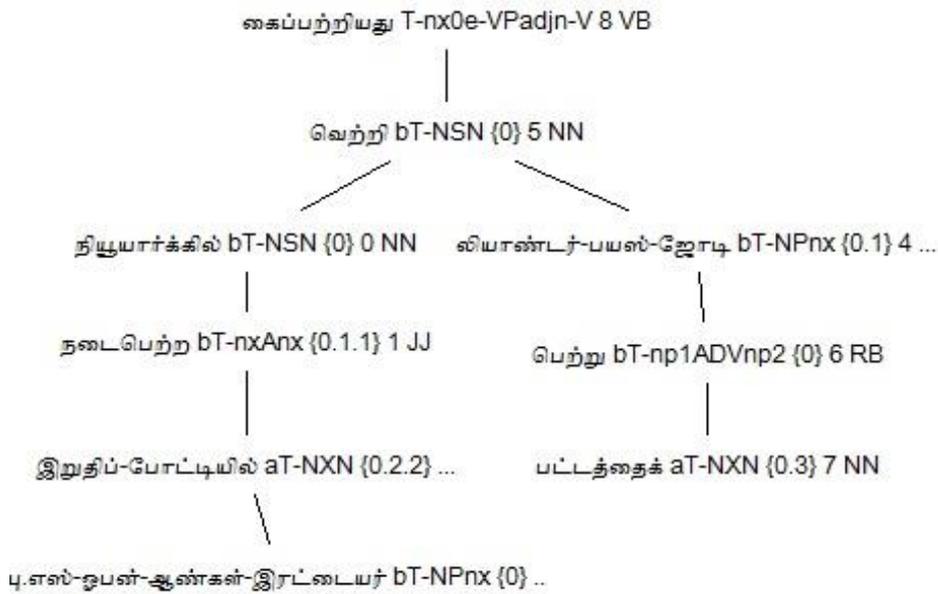

Derivation for Example 1

The derivation as above, clearly supports the 'root' verb (matrix verb, morph seperated). And the other dependencies such as nominal object and direct subject can also be seen here. Some non clausal adverbial dependencies can also be refined from this

***Example 2:*** *MOtirattai tiruṭiya vAliparai pOlIcAr tEṭi varukiṉṟaṉar*

This sentence has multiple parses maily due to a lexicosyntactic ambiguity. One prominent parse illustrated by Fig 8 and another parse illustrated by

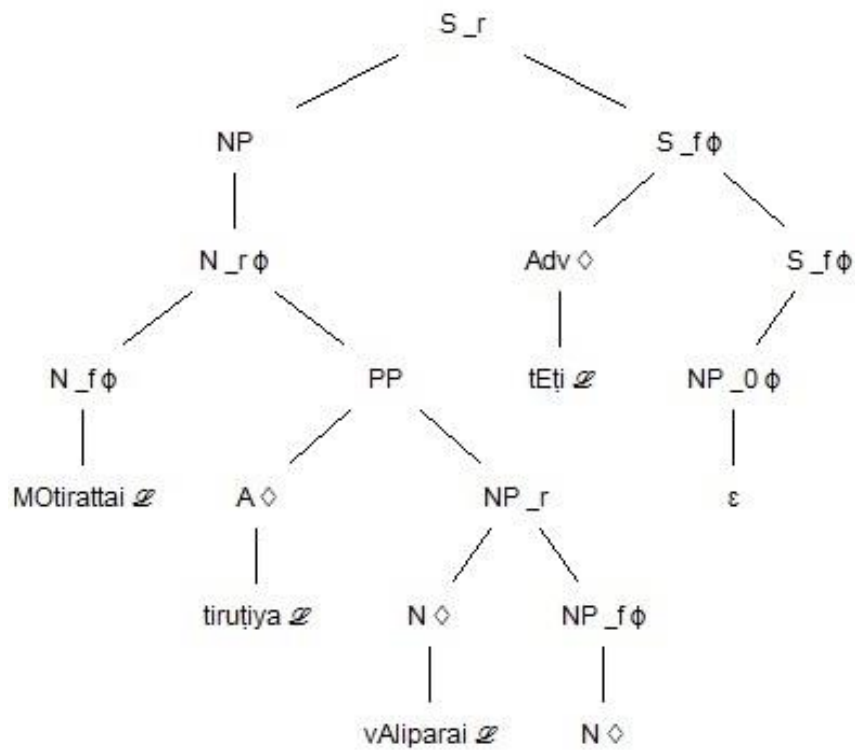

**Example2 :** *First Parse*
Derrived and Derivation Trees

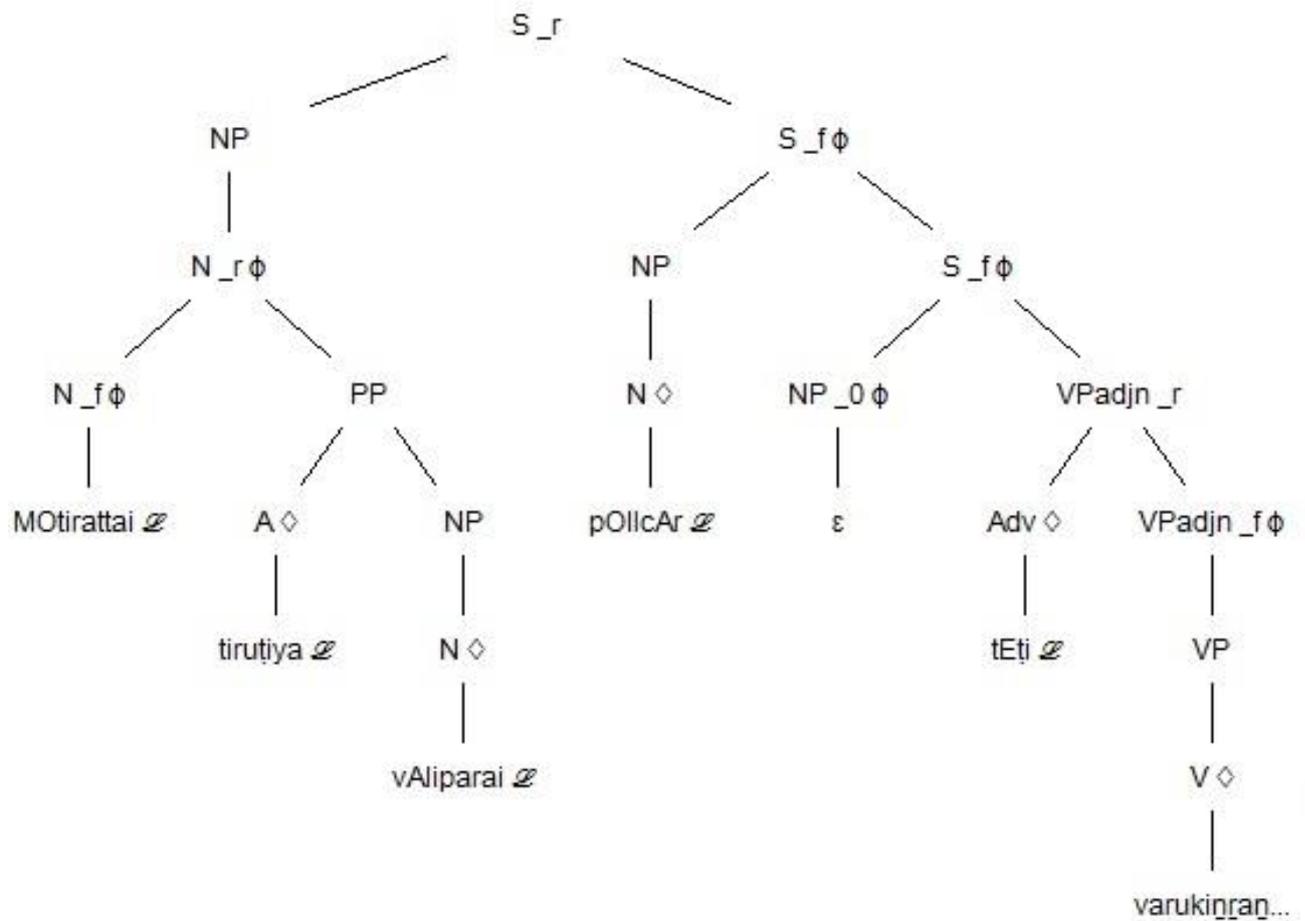

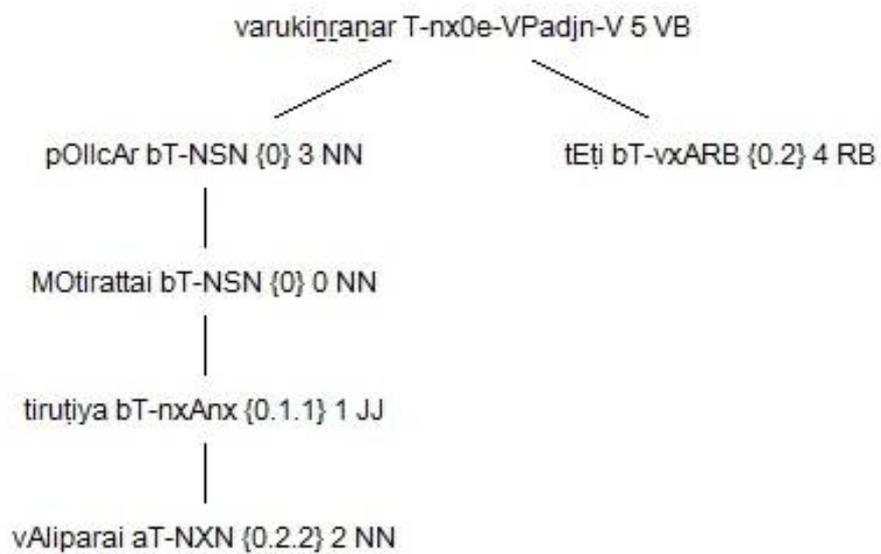

**Example 2:** Sencond Parse
Derived and Derivation Tree